\documentclass[10pt,twocolumn,letterpaper]{article}

\usepackage{cvpr}
\usepackage{times}
\usepackage{epsfig}
\usepackage{graphicx}
\usepackage{amsmath}
\usepackage{amssymb}
\usepackage{subfigure}
\usepackage{longtable}
\usepackage{multirow}
\usepackage{supertabular}
\usepackage{slashbox}
\usepackage{color}
\def\mR{\mathcal{R}}
\def\Red{\textcolor{red}}

\usepackage[pagebackref=true,breaklinks=true,bookmarks=false,colorlinks]{hyperref}

\cvprfinalcopy 

\def\cvprPaperID{436} 

\begin{document}

\title{Style Aggregated Network for Facial Landmark Detection}

\author{Xuanyi Dong$^{1}$, Yan Yan$^{1}$, Wanli Ouyang$^{2}$, Yi Yang$^{1}\thanks{Corresponding author.}$ \\
$^{1}$University of Technology Sydney, $^{2}$ The University of Sydney\\
{\tt\small \{xuanyi.dong,yan.yan-3\}@student.uts.edu.au;}\\
{\tt\small wanli.ouyang@sydney.edu.au; yi.yang@uts.edu.au}
}

\maketitle

\begin{abstract}

Recent advances in facial landmark detection achieve success by learning discriminative features from rich deformation of face shapes and poses.
Besides the variance of faces themselves, the intrinsic variance of image styles, e.g., grayscale vs. color images, light vs. dark, intense vs. dull, and so on, has constantly been overlooked.
This issue becomes inevitable as increasing web images are collected from various sources for training neural networks.
In this work, we propose a style-aggregated approach to deal with the large intrinsic variance of image styles for facial landmark detection.
Our method transforms original face images to style-aggregated images by a generative adversarial module.
The proposed scheme uses the style-aggregated image to maintain face images that are more robust to environmental changes.
Then the original face images accompanying with style-aggregated ones play a duet to train a landmark detector which is complementary to each other.
In this way, for each face, our method takes two images as input, i.e., one in its original style and the other in the aggregated style.
In experiments, we observe that the large variance of image styles would degenerate the performance of facial landmark detectors.
Moreover, we show the robustness of our method to the large variance of image styles by comparing to a variant of our approach, in which the generative adversarial module is removed, and no style-aggregated images are used.
Our approach is demonstrated to perform well when compared with state-of-the-art algorithms on benchmark datasets AFLW and 300-W.
Code is publicly available on GitHub: \url{https://github.com/D-X-Y/SAN}

\end{abstract}

\section{Introduction}

Facial landmark detection aims to detect the location of predefined facial landmarks, such as the corners of the eyes, eyebrows, the tip of the nose.
It has drawn much attention recently as it is a prerequisite in many computer vision applications.
For example, facial landmark detection can be applied to a large variety of tasks, including face recognition~\cite{zhucvpr2015high,liu2018exploring}, head pose estimation~\cite{wucvpr2017simultaneous}, facial reenactment~\cite{thies2016face2face} and 3D face reconstruction~\cite{liu2016joint}, to name a few.

\begin{figure}[t]
\center
\includegraphics[width=\columnwidth]{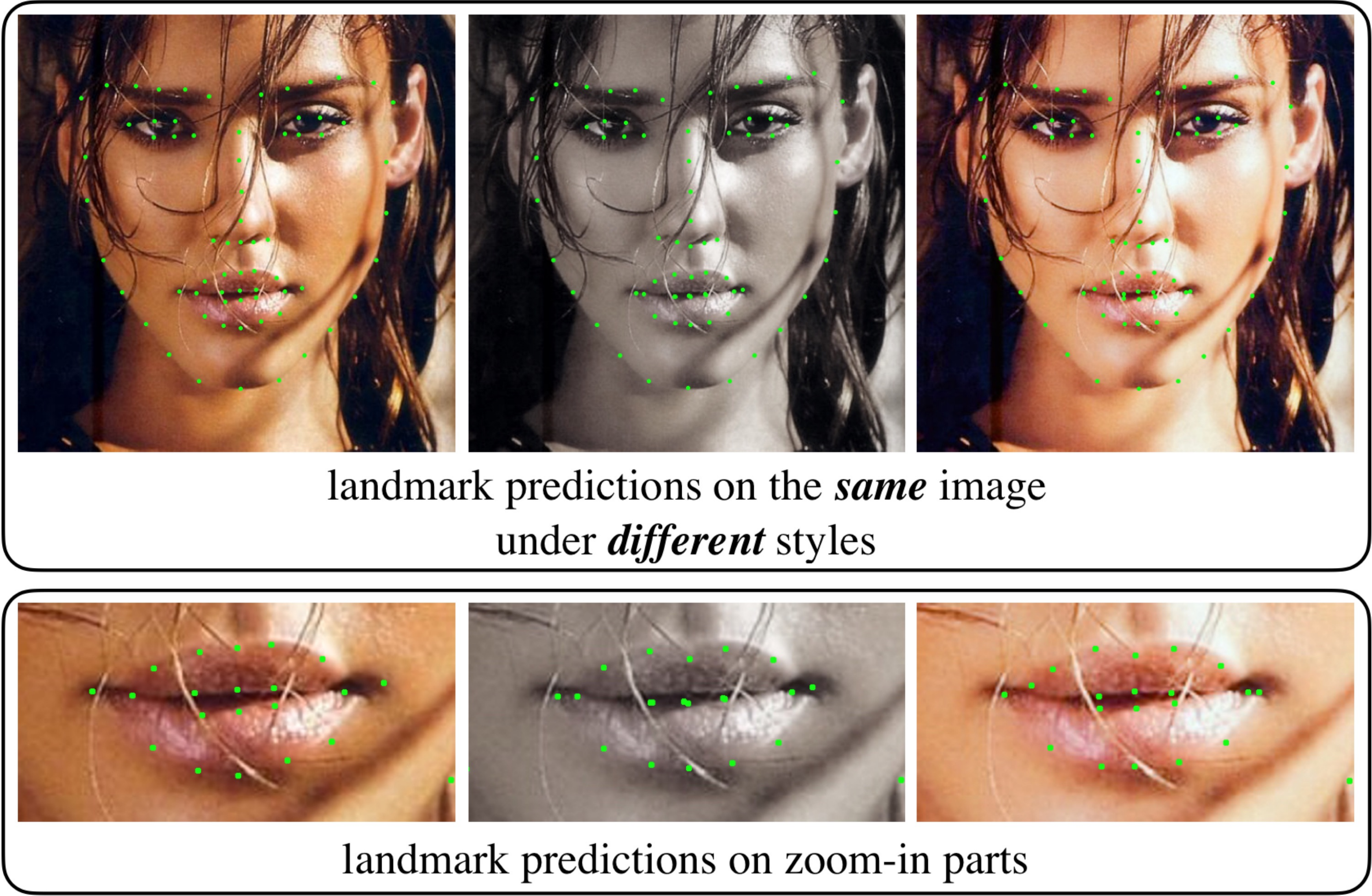}
\caption{
A face image in \emph{three} different styles and the locations of the facial landmarks predicted by a facial landmark detector on them.
The image styles, e.g., grayscale vs. color images, light vs. dark, intense vs. dull, can be quite distinct owing to various collection sources.
The contents of the above three images are identical. The only difference is the image style.
We apply a well-trained facial landmark detector to localize the facial landmarks.
The zoom-in parts show the deviation among the predicted locations of the same facial landmarks on different styled images.
}
\label{fig:high-level}
\end{figure}

Recent advances in facial landmark detection mainly focus on learning discriminative features from abundant deformation of face shapes and poses, different expressions, partial occlusions, and others~\cite{wucvpr2017simultaneous,zhu2016unconstrained,wucvpr2016constrained,jourabloo2017pose}.
A very typical framework is to construct features to depict the facial appearance and shape information by the convolutional neural networks (ConvNets) or hand-crafted features, and then learn a model, i.e., a regressor, to map the features to the landmark locations~\cite{xiong2013supervised,dollar2010cascaded,cao2014face,ren2014face,zhu2015face,zhang2014coarse,peng2016recurrent}.
Most of them apply a cascade strategy to concatenate prediction modules and update the predicted locations of landmarks progressively~\cite{zhang2014coarse,dollar2010cascaded,zhu2016unconstrained}.

\begin{figure}[t]
\center
\includegraphics[width=\columnwidth]{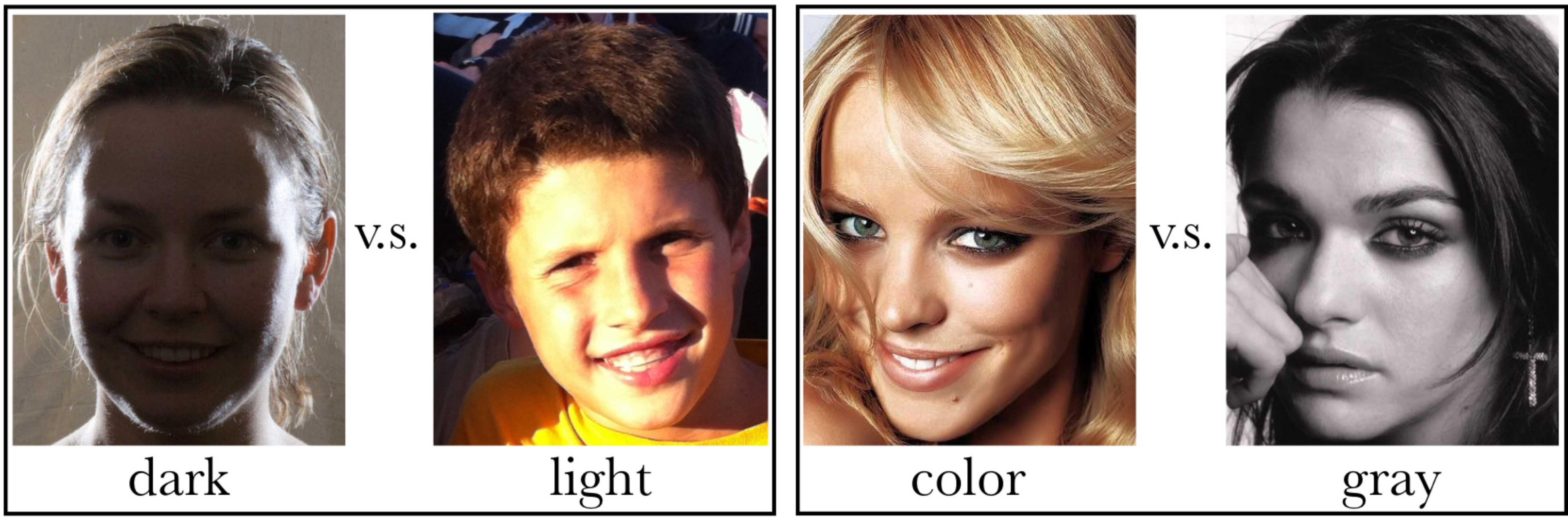}
\caption{
Face samples from 300-W dataset. Different faces have different styles, whereas the style information may not be approachable in most facial landmark detection datasets.
}
\label{fig:examples}
\end{figure}

However, the issue from image style variation has been overlooked by recent studies on facial landmark detection.
In real-world applications, face images collected in the wild usually are additionally under unconstrained variations~\cite{sagonas2013300,zhu2016unconstrained}.
Large intrinsic variance of image styles, e.g., grayscale vs. color images, light vs. dark, intense vs. dull, is introduced when face images are collected under different environments and camera settings.
The variation in image style causes the variation in prediction results.
For example, Figure~\ref{fig:high-level} shows \emph{three} different styles of a face image and the facial landmark predictions on them when applying a well-trained detector.
The contents of the three images are the same, but the visual styles are quite distinct, including original, grayscale and light.
We can observe that the location predictions of a same facial landmark on them can be different.
The zoom-in parts show the detailed deviation among the predicted locations of the same facial landmark on different styled images.
This intrinsic variance of image styles would distort the prediction of the facial landmark detector and further degenerate the accuracy, which will be empirically demonstrated later.
This problem commonly exists in the face in-the-wild landmark detection datasets~\cite{koestinger2011annotated,sagonas2013300}~(see Figure~\ref{fig:examples}), and becomes inevitable for such face images captured under uncontrolled conditions.

Motivated by the issue of large variance of different image styles, we propose a Style-Aggregated Network (SAN) for facial landmark detection, which is insensitive to the large variance of image styles.
The key idea of SAN is to first generate a pool of style-aggregated face images by the generative adversarial network (GAN)~\cite{goodfellow2014generative}.
Then SAN exploits the complementary information from both the original images and the style-aggregated ones.
The original images contain undistorted appearance contents of faces but may vary in image styles.
The style-aggregated images contain stationary environments around faces, but may lack certain shape information due to the less fidelity caused by GAN.
Therefore, our SAN takes both the original and style-aggregated faces together as complementary input, and applies a cascade strategy to generate the heatmap predictions which can be robust to the large variance of image styles.

To summarize, our contributions include:

\begin{enumerate}
\item To the best of our knowledge, we are the first to explicitly handle the problem caused by the variation of image styles in facial landmark detection problems, which has been overlooked in recent studies.
We further empirically verify the performance degeneration caused by the large variance of image styles.
    
\item To facilitate style analysis, we release two new facial landmark detection datasets, 300W-Styles ($\approx$~12000 images) and AFLW-Styles ($\approx$ 80000 images), by transferring the 300-W~\cite{sagonas2013300} and AFLW~\cite{koestinger2011annotated} into different styles.

\item We design a ConvNets architecture, i.e., Style-Aggregated Network (SAN), which exploits the mutual benefits of genuine appearance contents of faces and stationary environments around faces by simultaneously taking both original face images and style-unified ones.

\item In empirical studies, we verify the observation that the large variance of image styles would degenerate the performance of facial landmark detectors. 
Moreover, we show the insensitivity of SAN to the large variance of image styles and the state-of-the-art performance of SAN on benchmark datasets.
\end{enumerate}

\section{Related Work}

\subsection{Facial Landmark Detection}

Increasing researchers focus on facial landmark detection~\cite{sagonas2013300}.
The goal of facial landmark detection is to detect key-points in human faces, e.g., the tip of the nose, eyebrows, the eye corner and the mouth.
Facial landmark detection is a prerequisite for a variety of computer vision applications.
For example, Zhu~et~al.~\cite{zhucvpr2015high} take facial landmark detection results as input of 3D Morphable model.
Wu~et~al.~\cite{wucvpr2017simultaneous} propose a unified framework to deal with facial landmark detection, head pose estimation, and facial deformation analysis simultaneously, which couples each other.
Thies~et~al.~\cite{thies2016face2face} use facial landmark detection confidences of keypoints in feature alignment for facial reenactment.
Therefore, it is important to predict precise and accurate locations of the facial landmark.

\begin{figure*}[t]
\center
\includegraphics[width=0.96\textwidth]{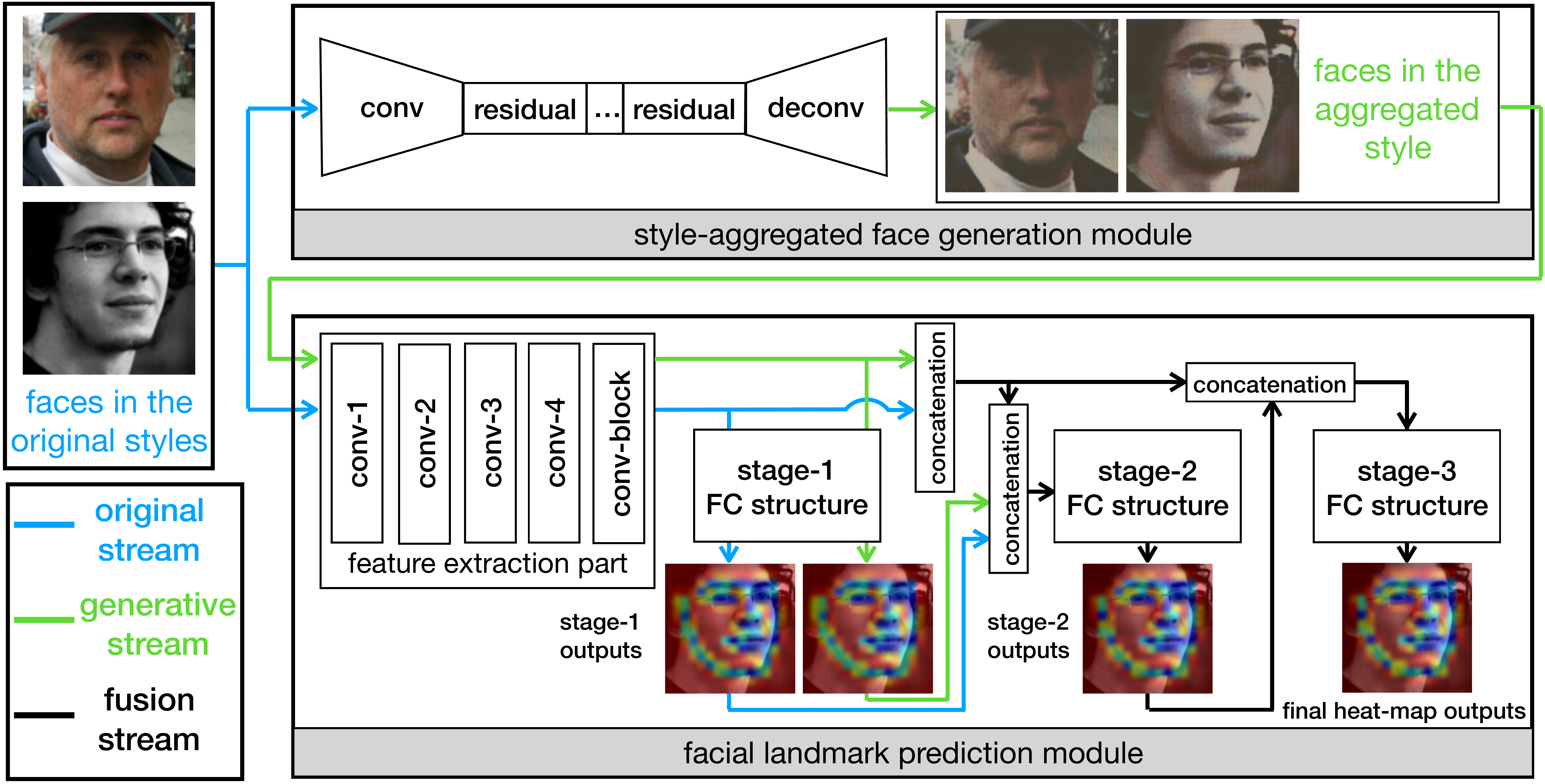}
\caption{
Overview of the SAN architecture.
Our network consists of two components.
The first is the style-aggregated face generation module, which transforms the input image into different styles and then combines them into a style-aggregated face.
The second is the facial landmark prediction module.
This module takes both the original image and the style-aggregated one as input to obtain two complementary features and then fuses the two features to generate heat-map predictions in a cascaded manner. ``FC'' means fully-convolution.
}
\label{fig:network}
\end{figure*}

A common approach to facial landmark detection problem is to learn a regression model~\cite{lv2017deep,xiong2013supervised,zhu2012face,bulat2017binarized,zhu2016unconstrained,cao2014face,xing2014towards}.
Many of them leverage deep CNN to learn facial features and regressors in an end-to-end fashion~\cite{sun2013deep,lv2017deep,zhu2016unconstrained}
with a cascade architecture to progressively update the landmark estimation~\cite{zhu2016unconstrained,sun2013deep,dollar2010cascaded}.
Yu~et~al.~\cite{yu2016deep} propose a deep deformation network to incorporates geometric constraints within the CNN framework.
Zhu~et~al.~\cite{zhu2016unconstrained} leverage cascaded regressors to handle extreme head poses and rich shape deformation.
Zhu~et~al.~\cite{zhu2015face} utilize a coarse search over a shape space with diverse shapes to overcome the poor initialization problem.
Lv~et~al.~\cite{lv2017deep} present a deep regression architecture with two-stage reinitialization to explicitly deal with the initialization problem.

Another category of facial landmark detection methods takes the advantages of end-to-end training from deep CNN model to learn robust heatmap for facial landmark detection~\cite{li2016face,wei2016convolutional,bulat2017far,bulat2016convolutional}.
Wei~et~al.~\cite{li2016face} and Newell~et~al.~\cite{newell2016stacked} take the location with the highest response on the heatmap as the coordinate of the corresponding landmarks.
Li~et~al.~\cite{li2016face} enhance the facial landmark detection by multi-task learning.
Bulat~et~al.~\cite{bulat2017far} propose a robust network structure utilizing the state-of-the-art residual architectures.

These existing facial landmark detection algorithms usually focus on the facial shape information, e.g., the extreme head pose~\cite{jourabloo2017pose} or rich facial deformation~\cite{zhu2016unconstrained}.
However, few of them engage in a consideration of the intrinsic variance of image styles, e.g., grayscale vs. color images, light vs. dark and intense vs. dull.
We also empirically demonstrate the performance fall caused by such intrinsic variance of image styles.
This issue has been overlooked by recent studies but becomes inevitable as increasing web images are collected from various sources.
Therefore, it is necessary to investigate the approach to dealing with the style variance, which is the focus of this paper.

Some researchers extend the landmark detection in the image to video settings~\cite{khan2017synergy,dong2018sbr,peng2016recurrent} or 3D settings~\cite{bulat2017far,simon2017kronecker}.
In contrast, we focuses on image-based landmark detection.

\subsection{Generative Adversarial Networks}

We leverage the generator of trained GAN to generate faces into different styles to combat the large variance of face image styles.

GANs are first proposed in~\cite{goodfellow2014generative} to estimate generative models via an adversarial process.
Following that, many researchers devoted great efforts to improve this research topic regarding theory~\cite{arjovsky2017wasserstein,chen2016infogan,li2017dualing,nowozin2016f,tolstikhin2017adagan} and applications~\cite{osokin2017gans,radford2015unsupervised,sixt2016rendergan,CycleGAN2017}.
Some of them contribute to face applications, such as makeup-invariant face verification~\cite{li2017anti} and face aging~\cite{antipov2017face}.
In this work, we leverage a recently proposed technique, CycleGAN~\cite{CycleGAN2017}, to integrate a face generation model in our detection network.
There are two different main focuses between this work and the previous works.
First, we aim to group images into specific styles in an unsupervised manner, while they usually assume a stationary style in a dataset.
Second, sophisticated face generation methods are not our target.

\section{Methodology}

How to design a neural network that is insensitive to the style variations for facial landmark detection?
As illustrated in Figure~\ref{fig:network}, we design a network by combine two sub-modules to solve this problem:
(1) The face generation module learns a neutral style of face images to combat the effect of style variations, i.e., transform faces with different styles into an aggregated style.
(2) The landmark prediction module leverages the complementary information from the neutral face and the original face to jointly predict the final coordinate for each landmark.

\subsection{Style-Aggregated Face Generation Module}\label{sec:style-module}

\begin{figure}[t]
\center
\includegraphics[width=\columnwidth]{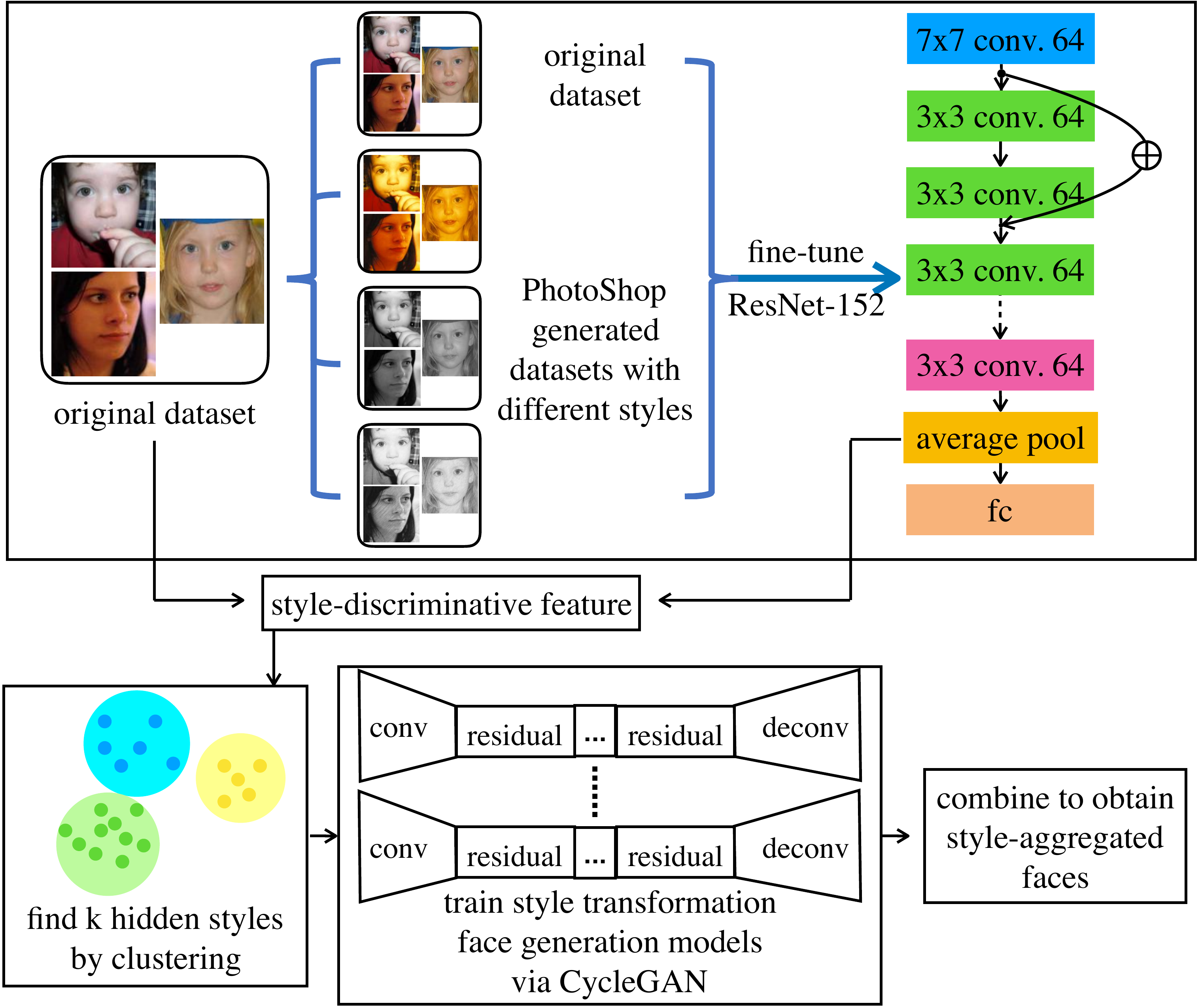}
\caption{
The pipeline to train the style-aggregated face generation module in an unsupervised way. We first utilize PS to transfer the original dataset into $C=3$ different styles. These transferred datasets accompanying with the original dataset are then used to fine-tune the ResNet-152 with $C+1$ classes.
The fine-tuned features from the global average pooling layer can be considered as the style-discriminative features.
We then leverage these features to cluster all images in the original dataset into $k$ clusters, which can potentially contain the information of hidden styles.
Lastly, we use these clustered data to train style transformation models via CycleGAN, and combine the trained models to obtain the final style-aggregated faces.
}
\vspace{-2mm}
\label{fig:pipeline}
\end{figure}

This module is motivated by the recent advances on image-to-image translation~\cite{isola2016image,CycleGAN2017} and style-transfer~\cite{dumoulin2016learned,gatys2016image,ulyanov2016texture}.
They can transform face images into a different style, whereas they require the style of images are already known in the training procedure as well as testing.
However, face images in facial landmark detection datasets are usually collected from multiple sources.
These images can have various styles, but we have no labels of these styles.
Therefore, current facial landmark datasets do not align with the settings of image-to-image translation, and can thus not directly apply their techniques to our problem.

We design an unsupervised approach to learn a face generation model to first transfer faces into different styles and then combine them into an aggregated style.
We first transfer the original dataset into three different styles by Adobe Photoshop (PS)
\footnote{Three styles: Light, Gray and Sketch. See details in Sec~\ref{sec:discussion}.}.
These three transferred datasets accompanying with the original dataset are regarded as four classes to fine-tune the classification model~\cite{simonyan2014very,he2016deep,szegedy2015going,dong2017more,yan2016image,xie2017aggregated,huang2017densely}.
The fine-tuned feature of the average-pooling layer thus has the style-discriminative characteristic, because the style information is learned in the training procedure by machine-generated style supervision.

To learn the stylized face generation model, we need to obtain the style information.
For most face in-the-wild datasets, we can identify that faces have different styles.
Figure~\ref{fig:examples} illustrates some examples of faces in various styles from 300-W~\cite{sagonas2013300}.
However, it is hard to label such datasets with different styles due to two reasons:
(1) Some style definitions are ambiguity, e.g., a face with light style can also be classified as the color.
(2) It requires substantial labors to label the style information.
Therefore, we leverage the learned style-discriminative feature to automatically cluster the whole dataset into $k$ hidden styles by k-means.

Lastly, we regard the face images in different clusters as different hidden styles, and we then train face generation models to transfer styles via CycleGAN.
CycleGAN is capable of preserving the structure of the input image because its cycle consistency loss guarantees the reconstructed images will match closely to the input images.
The overall pipeline is illustrated in Figure~\ref{fig:pipeline}.
The final output is several face generation models that can transfer face images into different styles, and average the transferred faces into the style-aggregated ones.

\subsection{Facial Landmark Prediction Module}\label{sec:prediction-module}

The facial landmark prediction module leverages the mutual benefit of both the original images and the style-aggregated ones to overcome negative effects caused by style variations.
This module is illustrated in Figure~\ref{fig:network}, where the green stream indicates the style-aggregated face and the blue stream represents the faces in the original styles.
The blue stream contains undistorted appearance contents of faces but may vary in image styles.
The green stream contains stationary environments around faces, but may lack certain shape information due to the less fidelity caused by GAN.
By leveraging their complementary information, we can generate more robust predictions. 
The architecture is inspired by CPM~\cite{wei2016convolutional}.
We use the first four convolutional blocks from VGG-16~\cite{simonyan2015very} followed by two additional convolution layers as feature extraction part.
The feature extraction part takes the face image ${\bf I_{o}} \in \mR^{h\times w}$ in the original styles and the one ${\bf I_{s}} \in \mR^{h\times w}$ from the style-aggregated stream as input, where $w$ and $h$ represent the width and the height of image.
In this part, each of the first three convolution blocks is followed by one pooling layer.
It thus outputs the features ${\bf F} \in \mR^{C \times h' \times w'}$ with eight times down-sample size compared to the input image~${\bf I}$, where $(h',w')=(h/8,w/8)$ and $C$ is the channel of the last convolutional layer.
The output features from the original and the style-aggregated faces are represented as ${\bf F_{o}}$ and ${\bf F_{s}}$, respectively.
Three subsequent stages are used to produce 2D belief maps~\cite{wei2016convolutional}.
Each stage is a fully-convolution structure.
Its output tensor ${\bf H} \in \mR^{(K+1) \times h' \times w'}$ has the same spatial size of the input tensor, where $K$ indicates the number of landmarks.
The first stage takes ${\bf F_{o}}$ and ${\bf F_{s}}$ as inputs and generate the belief maps for each of them, ${\bf H}_{o}$ and ${\bf H}_{s}$.
The second stage $g_{2}$ takes the concatenation of ${\bf F_{o}}$, ${\bf F_{s}}$, ${\bf H}_{o}$ and ${\bf H}_{s}$ as inputs, and output the belief map for stage-2:
{
\begin{align}
g_{2}({\bf F}_{o}, {\bf F_{s}}, {\bf H}_{o}, {\bf H}_{s}) = {\bf H}_{2} .
\end{align}
}
\vspace{-5mm}

\noindent The last stage is similar to the second one, which can be formulated as follows:
{
\begin{align}
g_{3}({\bf F}_{o}, {\bf F_{s}}, {\bf H}_{2}) = {\bf H}_{3} .
\end{align}
}
\vspace{-5mm}

\noindent Following~\cite{newell2016stacked,wei2016convolutional}, we minimize the following loss functions for each face image during the training procedure:
{
\begin{align}
Loss = \sum_{i\in\{o,s,2,3\}}||{\bf H}_{i}-{\bf H}_{i}^{*}||_{F}^{2} ,
\end{align}
}
\vspace{-3mm}

\noindent where ${\bf H}^{*}$ represents the ideal belief map.

To generate the final landmark coordinates, we first up-sample the belief map ${\bf H}_{3}$ to the original image size using bicubic interpolation.
We then use the argmax function on each belief map to obtain the coordinate of each landmark.

\section{Experiments}\label{sec:experiment}

\subsection{Datasets}\label{sec:datasets}

\textbf{300-W}~\cite{sagonas2013300}.
This dataset annotates five face datasets with 68 landmarks, LFPW~\cite{belhumeur2013localizing}, AFW~\cite{zhu2012face}, HELEN~\cite{le2012interactive}, XM2VTS, IBUG.
Following the common settings in~\cite{zhu2015face,lv2017deep}, we regard all the training samples from LFPW, HELEN and the full set of AFW as the training set, in which there is 3148 training images.
554 testing images from LFPW and HELEN form the common testing subset; 135 images from IBUG are regarded as the challenging testing subset.
Both of these two subsets form the full testing set.

\textbf{AFLW}~\cite{koestinger2011annotated}.
This dataset contains 21997 real-world images with 25993 faces in total.
They provide at most 21 landmark coordinates for each face but excluding invisible landmark.
Faces in AFLW usually have different pose, expression, occlusion or illumination, therefore causes difficulties to train a robust detector.
Following the same setting as in \cite{lv2017deep,zhu2016unconstrained}, we do not use the landmarks of two ears.
There are two types of AFLW splits, AFLW-Full and AFLW-Frontal following~\cite{zhu2016unconstrained}.
AFLW-Full contains 20000 training samples and 4386 testing samples.
AFLW-Front uses the same training samples as in AFLW-Full, but only use the 1165 samples with the frontal face as the testing set.

\subsection{Experiment Settings}\label{sec:setting}

\textbf{Training.}
We use PyTorch~\cite{paszke2017automatic} for all experiments.
To train the style-discriminative feature, we regard the original dataset and the PS-generated three datasets as four different classes.
We then use them to fine-tune ResNet-152 ImageNet pre-trained model, and we train the model with the learning rate of 0.01 for two epochs in total.
We use k-means to cluster the whole dataset into $k=3$ groups, and regard the group with the maximum element and the group with the minimum as two different style sets by default.
These two different groups are then used to train our style-unified face generation module via Cycle-GAN~\cite{CycleGAN2017}.
We follow the similar training settings as in~\cite{CycleGAN2017}, whereas we train our model with the batch size of 32 on two GPUs, and also set the identity loss in~\cite{CycleGAN2017} as 0.1.
To train the facial landmark prediction module, the first four convolutional blocks are initialized by VGG-16 ImageNet pre-trained model, and other layers are initialized using a Gaussian distribution with the variance of 0.01.
Lastly, we train the facial landmark prediction model with the batch size of 8 and weight decay of 0.0005 on two GPUs.
We start the learning rate at 0.00005 and reduce the learning rate at 30th/35th/40th/45th epochs by 0.5, and we then stop training at 50th epoch.
The face bounding box is expanded by the ratio of 0.2.
We use the random crop for pre-processing during training as data argumentation.

\begin{table}[t]
\small
\setlength{\tabcolsep}{3.2pt}
\centering
\begin{supertabular}{|c|c|c|c|} \hline
      Method                            & Common & Challenging & Full Set  \\\hline
      SDM \cite{xiong2013supervised}    & 5.57   & 15.40       & 7.52      \\
      ESR \cite{cao2014face}            & 5.28   & 17.00       & 7.58      \\
      LBF \cite{ren2016face}            & 4.95   & 11.98       & 6.32      \\
      CFSS \cite{zhu2015face}           & 4.73   & 9.98        & 5.76      \\
      MDM \cite{trigeorgis2016mnemonic} & 4.83   & 10.14       & 5.88      \\
      TCDCN \cite{zhang2014facial}      & 4.80   & 8.60        & 5.54      \\
      Two-Stage$_{OD}$ \cite{lv2017deep}& 4.36   & 7.56        & 4.99      \\
      Two-Stage$_{GT}$ \cite{lv2017deep}& 4.36   & 7.42        & 4.96      \\
      RDR \cite{xiao2017recurrent}      & 5.03   & 8.95        & 5.80      \\ 
 Pose-Invariant\cite{jourabloo2017pose} & 5.43   & 9.88        & 6.30      \\\hline
      SAN$_{OD}$                        & 3.41   & 7.55        & 4.24      \\
     {\bf SAN$_{GT}$}                & {\bf 3.34}& {\bf 6.60}  &{\bf 3.98} \\
\hline
\end{supertabular}
\vspace{2mm}
\caption{Normalized mean errors (NME) on 300-W dataset.
}
\vspace{-3mm}
\label{table:300W-ALL}
\end{table}

\begin{table*}[t]
\small
\setlength{\tabcolsep}{3.2pt}
\centering
\begin{supertabular}{|c|c c c c c c |c|} \hline
Methods   & SDM~\cite{xiong2013supervised} & ERT~\cite{kazemi2014one} & LBF~\cite{ren2016face} & CFSS~\cite{zhu2015face} & CCL~\cite{zhu2016unconstrained} & Two-Stage~\cite{lv2017deep} & SAN     \\ \hline
AFLW-Full & 4.05                           & 4.35                     & 4.25                   & 3.92                    & 2.72                       & 2.17                        & 1.91    \\ \hline
AFLW-Front& 2.94                           & 2.75                     & 2.74                   & 2.68                    & 2.17                       &  -                          & 1.85    \\\hline
\end{supertabular}
\vspace{2mm}
\caption{Comparisons of normalized mean (NME) errors on AFLW dataset.
}
\vspace{-2mm}
\label{table:aflw}
\end{table*}

\textbf{Evaluation.}
Normalized Mean Error (NME) is usually applied to evaluate the performance for facial landmark predictions~\cite{lv2017deep,ren2016face,zhu2016unconstrained}.
For 300-W dataset, we use the interocular distance to normalize mean error following the same setting as in~\cite{sagonas2013300,lv2017deep,cao2014face,ren2016face}.
For AFLW dataset, we use the face size to normalize mean error~\cite{lv2017deep}.
We also use Cumulative Error Distribution (CED) curve to compare the algorithms provided in~\cite{sagonas2016300}.
Area Under the Curve (AUC) @ 0.08 error is also employed for evaluation~\cite{bulat2017far,trigeorgis2016mnemonic}.

\subsection{Comparison with State-of-the-art Methods}\label{sec:compare}

{\bf Results on 300-W.}
Table~\ref{table:300W-ALL} shows the performance of different facial landmark detection algorithms on the 300-W.
We compare our approach with recently proposed state-of-the-art algorithms~\cite{lv2017deep,xiao2017recurrent,jourabloo2017pose}.
We compare our approaches based on two types of face bounding boxes:
(1) ground truth bounding box, denoted as GT; (2) official detector, denoted as OD.
SAN achieves very competitive results compared with others by using the same face bounding box (OD).
We improve the performance of NME on 300-W common set by relative 21.8\% compared to the state-of-the-art method.
It can further enhance our approach by applying a better initialization (GT).
This implies that SAN has potential to be more robust by incorporating the face alignment~\cite{lv2017deep} or landmark refinement~\cite{zhu2016unconstrained,trigeorgis2016mnemonic} methods.

{\bf Results on AFLW.}
We use the training/testing splits and the bounding box provided from~\cite{zhu2016unconstrained,zhu2015face}.
Table~\ref{table:aflw} shows the performance comparison on AFLW.
Our SAN also achieves the very competitive NME results, which are better than the previous state-of-the-art by more than 11\% on AFLW-Full.
On the AFLW-Front testing set, our result is also better than state-of-the-art by more than 14\%.
We find that more clusters and more generation models in style-aggregated face generation module will obtain a similar result as $k=3$, we thus use the setting of $k=3$ by default.

\begin{figure}[t]
\center
\subfigure[300-W Common Testing Set]{
\label{subfig:common}
\includegraphics[width=0.47\columnwidth]{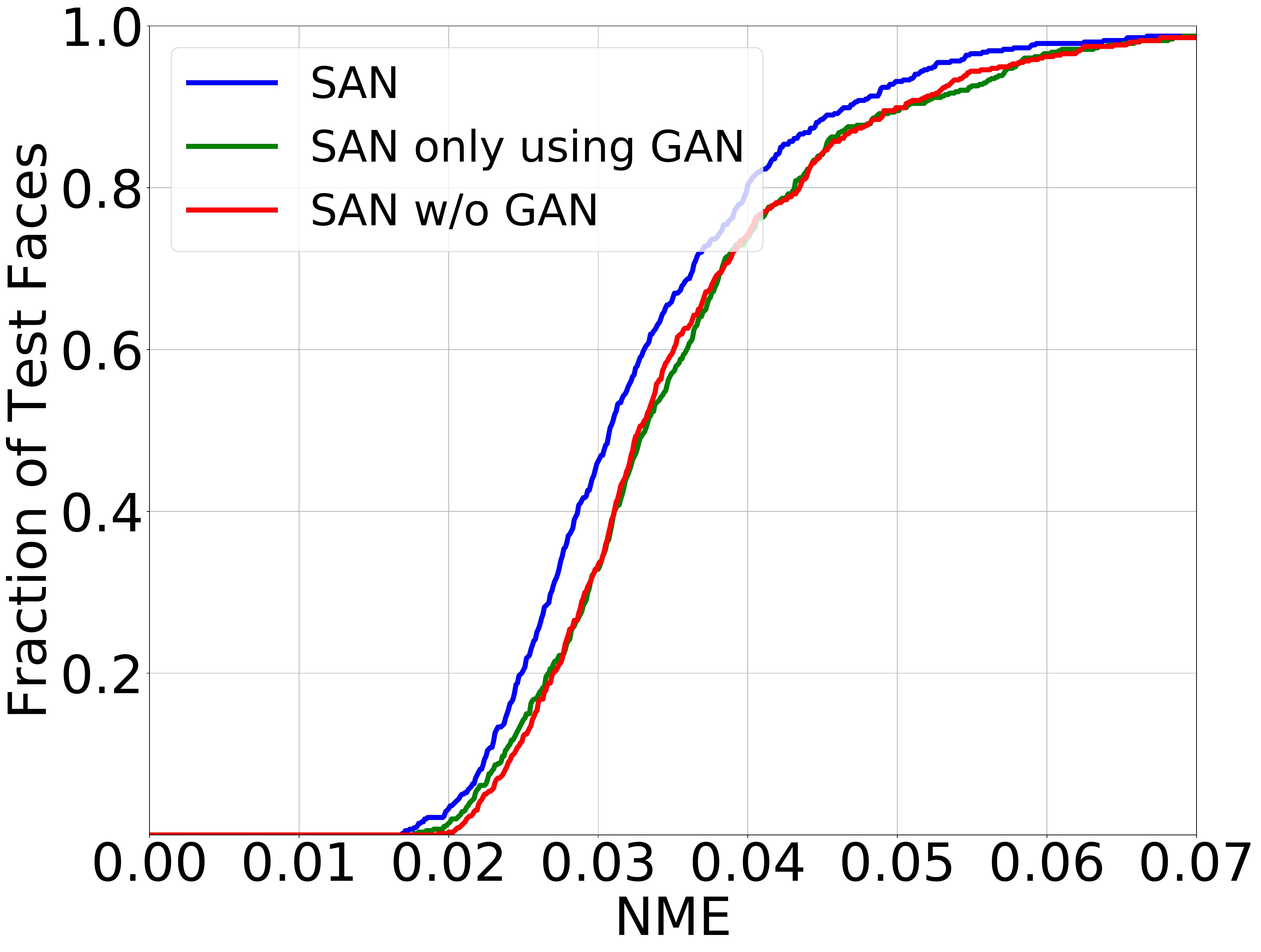}
}
\subfigure[300-W Challenging Testing Set]{
\label{subfig:challenging}
\includegraphics[width=0.47\columnwidth]{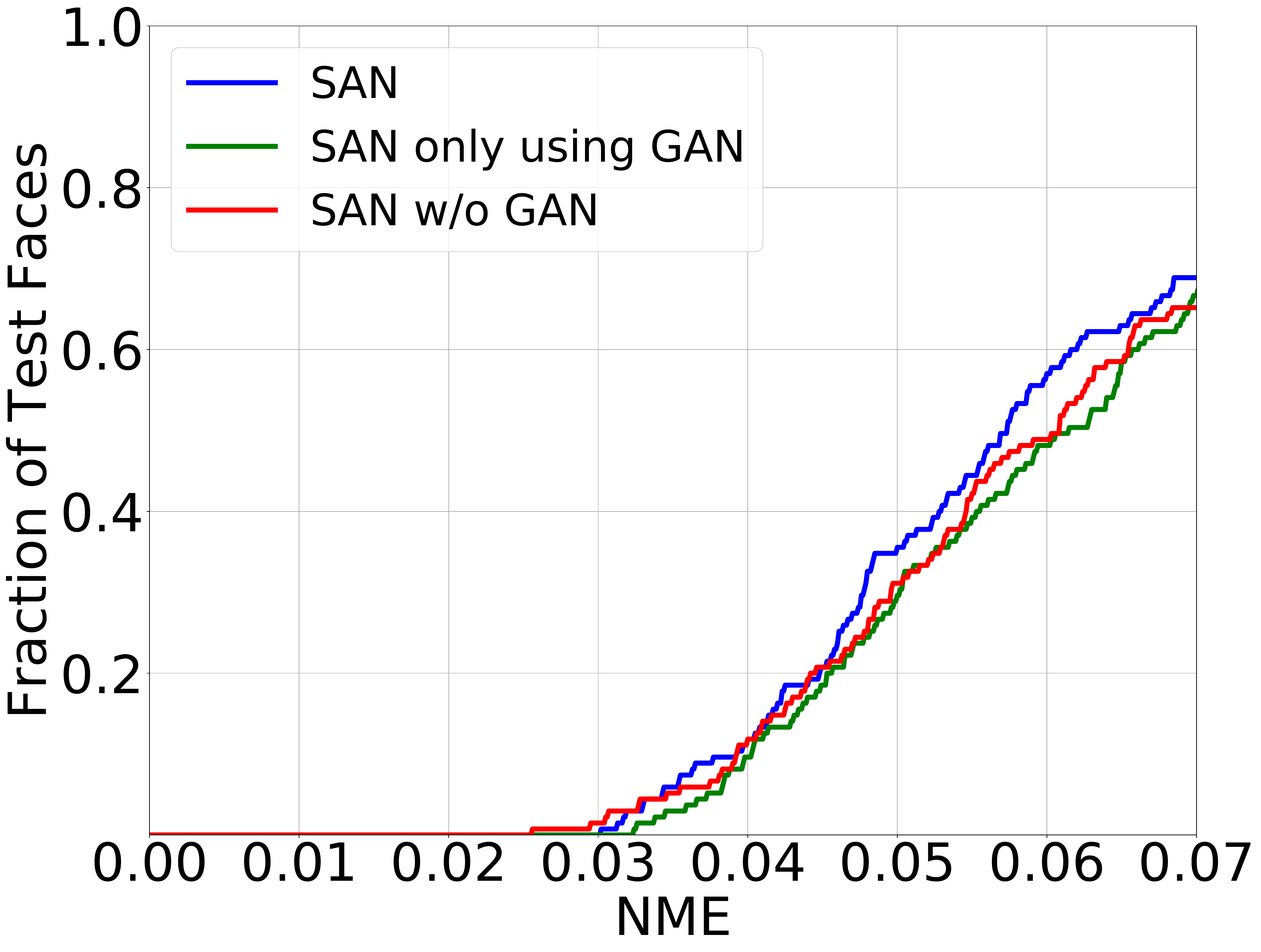}
}
\caption{CED curves for 300-W common and challenging testing sets.
The blue line shows the performance of SAN.
The green and red lines indicate SAN with only the style-aggregated face and with only the original face being the input, respectively.
}
\label{fig:pck_curve}
\end{figure}

SAN achieves new state-of-the-art results on two benchmark datasets, e.g., 300-W and AFLW.
It takes two complementary images to generate predictions which are insensitive to style variations.
The idea of using the two-stream input for facial landmark detection can be complementary to other algorithms~\cite{jourabloo2017pose,lv2017deep,xiao2017recurrent,zhu2016unconstrained}.
They usually do not consider the effect of image style, while the style-aggregated face in the two-steam input can handle this problem.

\subsection{Ablation Studies}\label{sec:ablation}

In this section, we first verify the significance of each component in our proposed SAN.
Figure~\ref{fig:pck_curve} shows the comparison regarding CED curves for our SAN and two variants of SAN on the 300-W common and testing sets.
As we can observe, the performance will significantly be deteriorated if we remove the original face image or the generated style-aggregated face image.
This observation demonstrates that taking two complementary face images as the input benefits the facial landmark prediction results.

\begin{figure}[t]
\center
\includegraphics[width=\columnwidth]{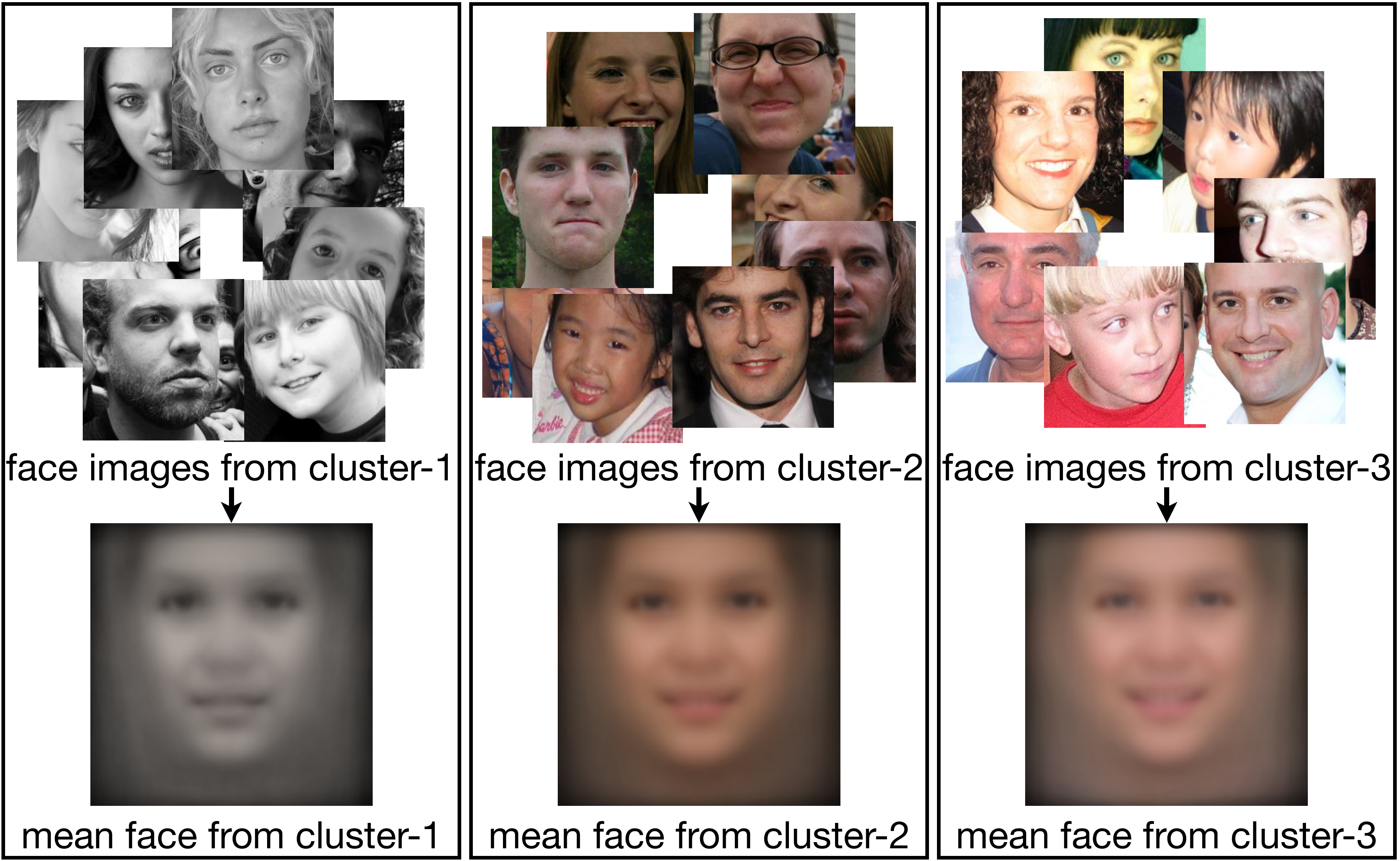}
\caption{
Qualitative results of the clustered face images from 300-W by using the style-discriminative features.
The face images in each cluster have some different hidden styles.
For example, the first cluster has many grayscale faces; the second cluster shows the dark illumination; the last cluster shows the light illumination.
We generate the mean face for each cluster.
These mean face images show the very \textit{similar} face, while they have quite \textit{different} environments.
}
\vspace{-2mm}
\label{fig:mean-face}
\end{figure}

Figure~\ref{fig:mean-face} shows the results of k-means clustering on 300-W dataset.
300-W dataset is the face in-the-wild dataset, where face images have large style variations but this style information is not approachable.
Our style-discriminative feature is capable of distinguishing images with different hidden styles.
We can find that most of the face images in one cluster share a similar style.
The mean face images generated from three clusters contain different styles.
If we directly use ImageNet pre-trained features for k-means clustering, we can not guarantee to group faces into different hidden styles.
In experiments, we find that ImageNet pre-trained features tend to group face images by the gender or other information.

\subsection{Discussions of Benchmark Datasets}\label{sec:discussion}

\begin{figure}[t]
\center
\includegraphics[width=\linewidth]{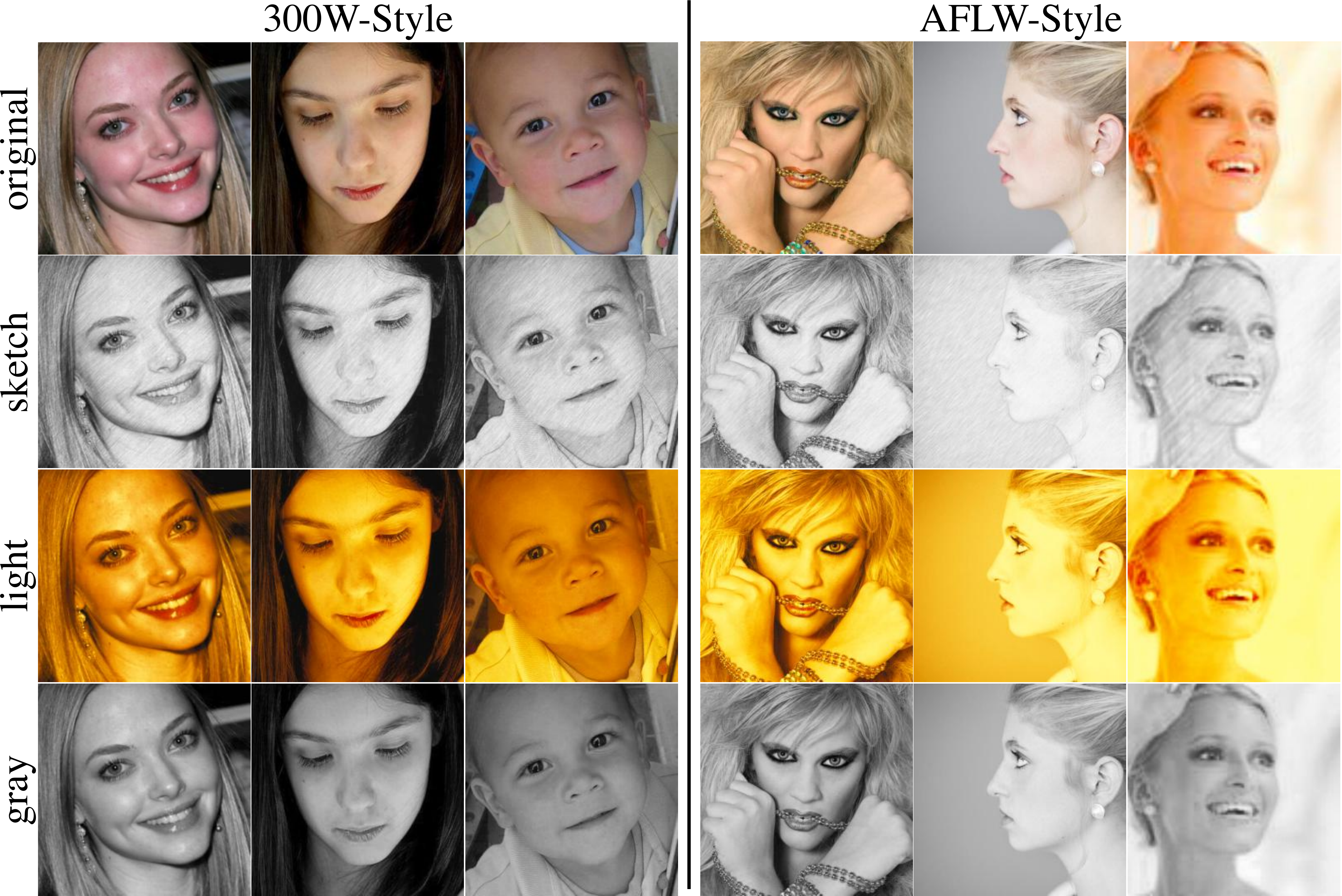}
\caption{
Our PS-generated datasets based on 300-W and AFLW with the original and three synthetic styles, i.e., sketch, light and gray.
These datasets have different styles and can be used to facilitate style analysis.
}
\vspace{-2mm}
\label{fig:dataset}
\end{figure}

Facial landmark detection datasets with constrained face images~\cite{milborrow2010muct} usually have the similar environment for each image. There are only small style changes in these datasets, and they may also not be applicable for real-world applications due to the small face variance. We thus do not discuss these datasets in this paper.
The face in-the-wild datasets~\cite{sagonas2013300,koestinger2011annotated} contain face images with large intrinsic variance. However, this intrinsic variance information is not available from the official datasets, but can also affect the predictions of the detector.
Therefore, we propose two new datasets, 300W-Style and AFLW-Style, to facilitate the style analysis for facial landmark detection problem.

\begin{table}[b]
\small
\setlength{\tabcolsep}{3.2pt}
\centering
\begin{supertabular}{|c|c|c|c|c|}\hline 
\backslashbox{Test}{Train} &   Original       &   Light                 &  Gray                    &  Sketch  \\\hline\hline
                  \multicolumn{5}{|c|}{SAN w/o GAN}       \\\hline
Original                   &   3.37           &    3.56                 &  3.77                    &   3.92   \\\hline 
Light                      &   3.61           &    3.41                 &  4.01                    &   4.13   \\\hline 
Gray                       &   3.47           &    3.79                 &  3.43                    &   3.60   \\\hline
Sketch                     &   3.71           &    3.97                 &  3.66                    &   3.40   \\\hline\hline
                  \multicolumn{5}{|c|}{SAN}                    \\\hline
\multirow{2}{*}{Original}  &   3.34           &    3.44                 &  3.46                    &   3.54        \\
                     &($\Red{\uparrow0.8\%}$) & ($\Red{\uparrow3.3\%}$) & ($\Red{\uparrow8.2\%}$)  & ($\Red{\uparrow9.7\%}$)\\\hline 
\multirow{2}{*}{Light}     &   3.48           &    3.39                 & 3.56                     &   3.68        \\
                     &($\Red{\uparrow3.6\%}$) & ($\Red{\uparrow0.5\%}$) & ($\Red{\uparrow11.2\%}$) & ($\Red{\uparrow10.9\%}$)\\\hline 
\multirow{2}{*}{Gray}      &   3.45           & 3.56                    & 3.38                     &   3.52      \\
                     &($\Red{\uparrow0.6\%}$) & ($\Red{\uparrow6.1\%}$) & ($\Red{\uparrow1.4\%}$)  & ($\Red{\uparrow2.2\%}$)\\\hline
\multirow{2}{*}{Sketch}    &   3.53           &    3.62                 &  3.55                    &   3.35      \\
                      &($\Red{\uparrow4.9\%}$)& ($\Red{\uparrow8.8\%}$) & ($\Red{\uparrow3.0\%}$)  & ($\Red{\uparrow1.4\%}$)\\\hline 
\end{supertabular}
\vspace{2mm}
\caption{
Comparisons of NME on the 300W-Style common testing set.
We use different styles for training and testing.
}
\vspace{-2mm}
\label{table:300W-Common}
\end{table}

As shown in Figure~\ref{fig:dataset}, 300W-Style consists of four different styles, original, sketch, light and gray.
The original part is the original 300-W datasets, and the other three synthetic styles are generated using PS.
Each image in 300W-Style is corresponding to one image in the 300-W dataset, and we thus directly use the annotation provided from 300-W for our 300W-Style.
AFLW-Style is similar as 300W-Style, which transfer the AFLW dataset into three different styles.
For training and testing split, we follow the common settings of the original datasets~\cite{sagonas2013300,koestinger2011annotated}.

{\bf Can PS-generated images be realistic?}
Internet users usually use PS (or similar software) to change image styles and/or edit image content; thus PS-generated images are indeed realistic in many real-world applications.
In addition, we have chosen three representative filters to generate images of different styles. These filters have been widely used by users to edit their photos and upload to the Internet. Therefore, the proposed datasets are realistic.

\begin{table}[t]
\small
\setlength{\tabcolsep}{3.2pt}
\centering
\begin{supertabular}{|c|c|c|c|c|}\hline 
\backslashbox{Test}{Train} &   Original       &   Light                 &  Gray                   & Sketch                   \\\hline\hline
                  \multicolumn{5}{|c|}{SAN w/o GAN}                                  \\\hline
Original                   &    6.88          &    7.82                 &   7.84                  &   7.74                   \\\hline 
Light                      &    7.31          &    7.16                 &   8.91                  &   8.67                   \\\hline 
Gray                       &    7.08          &    8.59                 &   6.77                  &   6.98                   \\\hline
Sketch                     &    7.59          &    8.68                 &   7.17                  &   6.83                   \\\hline\hline
                  \multicolumn{5}{|c|}{SAN}                    \\\hline
\multirow{2}{*}{Original}  &    6.60          &    7.00                 &   6.73                  &   6.97                   \\
                      &($\Red{\uparrow4.1\%}$)&($\Red{\uparrow10.5\%}$) &($\Red{\uparrow14.2\%}$) &($\Red{\uparrow9.9\%}$)   \\\hline 
\multirow{2}{*}{Light}     &     7.15         &    7.08                 &   7.26                  &   7.15                   \\
                     &($\Red{\uparrow2.2\%}$) &($\Red{\uparrow1.1\%}$)  &($\Red{\uparrow18.5\%}$) & ($\Red{\uparrow17.5\%}$) \\\hline 
\multirow{2}{*}{Gray}      &     6.91         &    7.18                 &   6.69                  &   6.97                   \\
                     &($\Red{\uparrow2.4\%}$) & ($\Red{\uparrow16.4\%}$)&($\Red{\uparrow1.1\%}$) & ($\Red{\uparrow0.2\%}$)  \\\hline
\multirow{2}{*}{Sketch}    &     7.08         &    7.64                 &   6.95                  &   6.77                   \\
                     &($\Red{\uparrow6.7\%}$) & ($\Red{\uparrow12.0\%}$)&($\Red{\uparrow3.1\%}$)  & ($\Red{\uparrow0.8\%}$)  \\\hline 
\end{supertabular}
\vspace{2mm}
\caption{
Comparisons of NME on the 300W-Style challenging testing set.
We use different styles for training and testing.
}
\vspace{-2mm}
\label{table:300W-Challenging}
\end{table}

\begin{table}[t]
\small
\setlength{\tabcolsep}{3.2pt}
\centering
\begin{supertabular}{|c|c|c|c|c|}\hline 
\backslashbox{Test}{Train} &   Original       &            Light        &          Gray           &   Sketch \\\hline\hline
                  \multicolumn{5}{|c|}{SAN w/o GAN}                                  \\\hline
Original                   &    4.06          &             4.39        &          4.57           &    4.67  \\\hline 
Light                      &    4.33          &             4.14        &          4.97           &    5.02  \\\hline 
Gray                       &    4.19          &             4.73        &          4.08           &    4.26  \\\hline
Sketch                     &    4.47          &             4.89        &          4.35           &    4.07  \\\hline\hline
                  \multicolumn{5}{|c|}{SAN}                                          \\\hline
\multirow{2}{*}{Original}  &    3.98          &             4.14        &          4.10           &    4.21  \\
					  &($\Red{\uparrow1.9\%}$)&($\Red{\uparrow5.7\%}$) &($\Red{\uparrow10.2\%}$) &($\Red{\uparrow9.9\%}$)   \\\hline 
\multirow{2}{*}{Light}     &    4.20          &             4.12        &          4.29           &    4.36  \\
                      &($\Red{\uparrow3.0\%}$)&($\Red{\uparrow0.4\%}$) &($\Red{\uparrow13.7\%}$) &($\Red{\uparrow13.1\%}$)  \\\hline 
\multirow{2}{*}{Gray}      &    4.13          &             4.27        &          4.03           &    4.20  \\
					  &($\Red{\uparrow1.4\%}$)&($\Red{\uparrow9.7\%}$) &($\Red{\uparrow1.2\%}$) &($\Red{\uparrow1.4\%}$)  \\\hline
\multirow{2}{*}{Sketch}    &    4.23          &             4.41        &          4.21           &    4.02  \\
				      &($\Red{\uparrow5.4\%}$)&($\Red{\uparrow6.7\%}$) &($\Red{\uparrow3.2\%}$) &($\Red{\uparrow1.2\%}$) \\\hline 
\end{supertabular}
\vspace{2mm}
\caption{
Comparisons of NME on the 300W-Style full testing set.
We use different styles for training and testing.
}
\vspace{-2mm}
\label{table:300W-Full}
\end{table}

\begin{figure*}[t]
\center
\includegraphics[width=\textwidth]{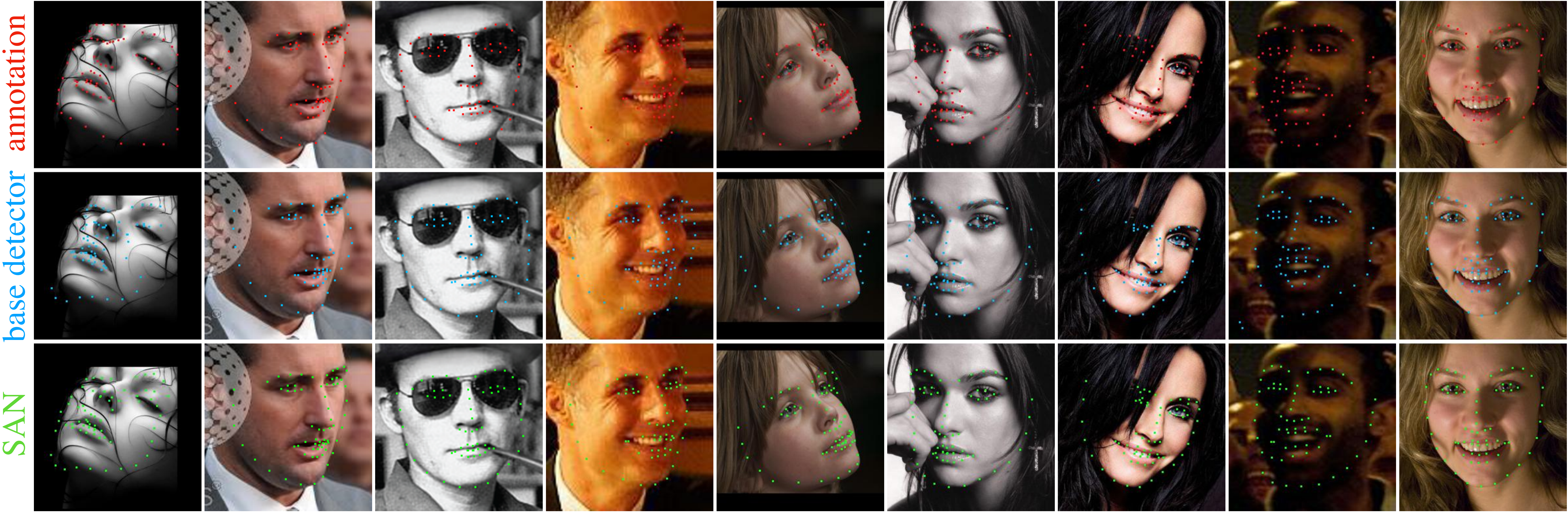}
\caption{Representative results on 300-W. The red points in the first line indicate the ground-truth landmarks.
The blue points in the second line and the green points in the third line indicate the landmark predictions from the base detector and SAN, respectively.
}
\vspace{-3mm}
\label{fig:results}
\end{figure*}

{\bf Effect of SAN for style variances.}
These two proposed datasets can be used to analyze the effect of face image styles for facial landmark detection.
We consider the situation that testing set has a different style with the training set.
For example, we train the detector on the light-style 300-W training set and evaluate the well-trained detector on 300-W testing sets with different styles.
Table~\ref{table:300W-Common}, Table~\ref{table:300W-Challenging} and Table~\ref{table:300W-Full} show the evaluation results of 16 training and testing style combinations, i.e., four different training styles multiply four different testing styles.
Our SAN algorithm is specifically designed to deal with style variances for face landmark detection.
When style variance between the training and testing sets is large (e.g., light and gray), our approach usually obtains a significant improvement.
However, if style variance between the training and testing sets is not that large (e.g., gray and sketch), the improvement of SAN is less significant.
On average, SAN obtains 7\% relative improvement on the full testing set of the 300W-Style dataset when the training style is different from the testing style.
Moreover, our SAN achieves consistent improvements over all the 16 different train-test style combinations.
This demonstrates the effectiveness of our method.

{\bf Self-Evaluation:}
We compare two variants of our SAN:
(1) train SAN without GAN using the training set of AFLW-Style and the testing set of AFLW.
This can be considered as data argumentation, because the amount of training data that we use is four times larger than the original one.
In this case, our SAN can achieve 79.82 AUC@0.08 on AFLW-Full by only using the original AFLW training set, while the data argumentation one achieves a worse performance, 78.99 AUC@0.08, than SAN.
SAN is better than the data argumentation way, which uses our PS-generated images as additional training data.
(2) replace the style-aggregated stream of SAN by a Photo-generated face image.
If we train the detector on the original style 300-W training set and test it on the gray style 300-W challenging test set, our SAN can achieve 6.91 NME.
However, replacing the style-aggregated stream by light style images can only achieve 7.30 NME, which is worse than ours.
SAN can always achieve better results than the replaced variant, except for replacing the style-aggregated stream by the testing style.
SAN can automatically learn the hidden styles in the dataset and generate the style-aggregated face images. This automatic way is better than providing images with a fixed style.

{\bf Error Analysis:}
The faces in uncontrolled conditions have large variations regarding the image style.
Detectors will usually fail when image style changes a lot, whereas our SAN is insensitive to this style change.
Figure~\ref{fig:results} shows the qualitative results of our SAN and the base detector on 300-W.
The first line shows the ground truth landmarks.
The second and third lines show the predictions from SAN without GAN and SAN, respectively.
In the first column, the base detector fails for the predictions on the face contour, while the predictions from SAN still preserves the overall structure.
In the fourth column, some perdition from the base detector drifts to the right, while SAN not.

\section{Conclusion \& Future Work}
The large intrinsic variance of image styles, which comes from their uncontrolled collection sources, has been overlooked by recent studies in facial landmark detection.
To deal with this issue, we propose a style-aggregated network (SAN).
SAN takes two complementary images for each face, one in the original style and the other in the aggregated style that is generated by GAN.
Empirical studies verify that style variations degenerate the performance of landmark detection, and SAN is robust to the large variance of image styles.
Additionally, SAN achieves state-of-the-art performance on 300-W and AFLW datasets.

The first step of SAN is to generate the style-aggregated images.
This step can be decoupled from our landmark detector, and potentially used to improve other landmark detectors~\cite{cao2014face,ren2016face,zhu2015face,zhang2014facial,ouyang2014multi}.
Moreover, the intrinsic variance of image styles also exists in other computer vision tasks, such as object detection~\cite{dong2017dual,ren2017faster,ouyang2012discriminative,dai2016r,liu2017recurrent} and person re-identification~\cite{wu2018exploit,zhao2013unsupervised,zhong2018camera,ma2017self}.
Therefore, the style-aggregation method can also be used to solve the problem of the style variance in other applications.
In our future work, we will explore how to generalize the style-aggregation method for other computer vision tasks.

{\bf Acknowledgment.} Yi Yang is the recipient of a Google Faculty Research Award.
Wanli Ouyang is supported by SenseTime Group Limited.
We acknowledge the Data to Decisions CRC (D2D CRC) and the Cooperative Research Centres Programme for funding this research.

{\small
\bibliographystyle{ieee}
\bibliography{egbib}
}

\end{document}